\documentclass[journal]{IEEEtran}
\pdfoutput=1 

\newcommand{\etal}{\textit{et al}. }
\usepackage{cite}
\usepackage{amsmath,amssymb,amsfonts}
\usepackage{algorithmic}
\usepackage{graphicx}
\usepackage{textcomp}
\usepackage{xcolor}
\usepackage{dblfloatfix}
\usepackage{authblk}

\usepackage[]{footmisc}

\usepackage{cuted}
\usepackage{capt-of}
\def\BibTeX{{\rm B\kern-.05em{\sc i\kern-.025em b}\kern-.08em
		T\kern-.1667em\lower.7ex\hbox{E}\kern-.125emX}}

\makeatletter

\ifCLASSINFOpdf
\else
\fi

\begin{document}
%
\title{Pipeline for 3D reconstruction of the human body from AR/VR headset mounted egocentric cameras}
%
%
%


\author{
\IEEEauthorblockN {
   Shivam~Grover $^{1,+}$,       
   Kshitij~Sidana $^{1,+}$       
   and Vanita~Jain $^{1,*}$ 
}

\IEEEauthorblockA {
  $^{1}$Bharati Vidyapeeth's College of Engineering, New Delhi, India\\
}
\IEEEauthorblockA {
  $^{+}$Contributed equally to this work as first authors.\\
}
\IEEEauthorblockA {
    $^{*}$Corresponding author: Vanita Jain vanita.jain@bharatividyapeeth.edu
}
}

\maketitle

\begin{abstract}
In this paper we propose a novel pipeline for the 3D reconstruction of the full body from egocentric viewpoints.
3-D reconstruction of the human body from egocentric viewpoints is a challenging task as the view is skewed and the body parts farther from the cameras are occluded. One such example is the view from cameras installed below VR headsets. To achieve this task, we first make use of conditional GANs to translate the egocentric views to full body third person views. This increases the comprehensibility of the image and caters to occlusions. The generated third person view is further sent through the 3D reconstruction module that generates a 3D mesh of the body. We also train a network that can take the third person full body view of the subject and generate the texture maps for applying on the mesh. The generated mesh has fairly realistic body proportions and is fully rigged allowing for further applications such as real time animation and pose transfer in games.
This approach can be key to new domain of mobile human telepresence.
\end{abstract}


%
\IEEEpeerreviewmaketitle

\section{Introduction}
%
%
%
%
\IEEEPARstart{T}{HE} process of 3D reconstruction refers to the construction of the mesh and the corresponding texture of an object from a 2D image of it. 3D reconstruction of the human body using wearable cameras (such as virtual reality headsets) has many exciting applications such as walk-in movies, interactive TV shows, virtual meetings, remote training, and the most anticipated of all, personalized gaming experiences. While recent works which do facial reconstructions \cite{Chen2013DeformableMF}, pose estimation \cite{xu2019mo2cap2}\cite{Rhodin2016EgoCapEM}, and environment reconstruction \cite{Cha2018TowardsFM} from VR (virtual reality) headsets cameras have shown some spectacular results, no success has been seen in reconstructing the 3D mesh of the full body using only images from head-worn cameras. In our work, we not only achieve the 3D reconstruction of the full body, but we also infer the full body textures of the body all from a single pair of front-back egocentric images only. We also show the reconstructed body in novel poses and viewpoint.

In general, the goal of image-based 3D reconstruction is to infer the 3D geometry and structure of objects and scenes from one or multiple 2D images. The field of 3D reconstruction from images has been widely explored and has produced remarkable methods to do it. These methods include stereo-based techniques, shape from silhouette, or shape by space carving methods, using multiple images of the same object captured by well-calibrated cameras. This is generally achieved by placing one or more fixed cameras and sensors around the subject. The cost of setup is high and it's feasibility is low as it requires dedicated sensors and a large space to be set up. Since it cannot be moved so easily, this lowers its portability and the mobility of the user which is a huge disadvantage for virtual reality based applications. In developing countries like India where more than 75\% of the population lives in a house with less than 2 rooms, being able to afford and accommodate such a setup is like a dream.

We envision a future where to be able to achieve a sense of virtual physical presence of your peers, all you need is a head-wear gear. Virtual reality based applications have seen a rise in popularity in the last decade. With the work-from-home culture being at its peak, there is also a great demand and popularity for telepresence,  virtual meetings and conferences.

\begin{figure}[t]
	\centering
	\includegraphics[width=\linewidth]{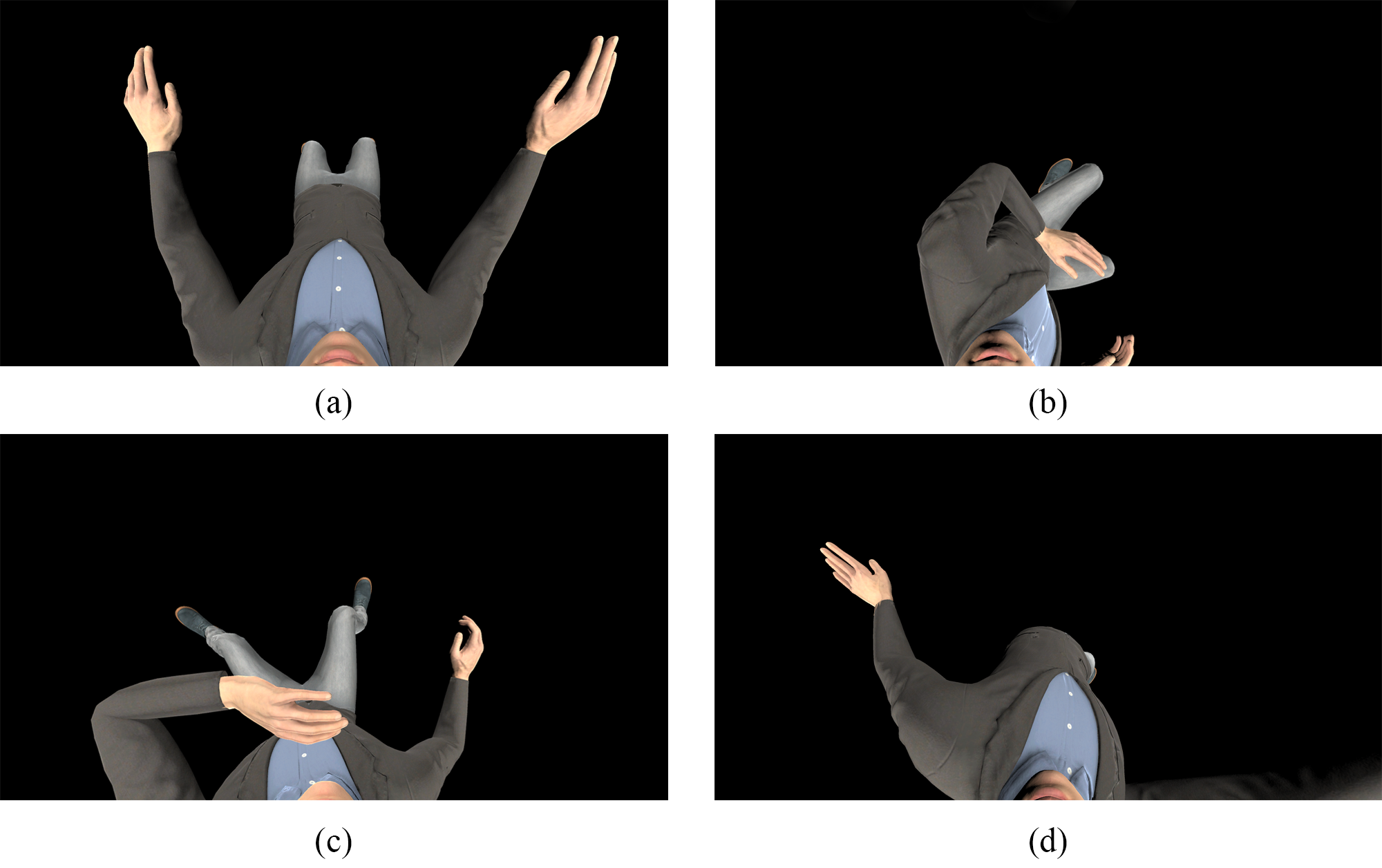}
	\caption{Sample images from a camera placed under a VR headset capturing the front body. In (a), it can be seen that even in an image where the person is standing straight, there is severe distortion due to perspective. In (b) and (c) some portion of the body is occluded by the hands. In (d), the person is bending slightly backwards which causes the legs to be occluded.}
	\label{fig:occlusion}
\end{figure}

While 3D reconstruction from static cameras and cameras capturing the subject from a distance is a widely explored area, 3D reconstruction from egocentric cameras is relatively new. There are certain aspects of egocentric images that pose a problem in working with them. The view of the body is not optimal for full body reconstruction since the view is massively distorted due to perspective, most of the body is occluded and while bending backwards, a major portion of the body goes out of the field of view of the camera (Fig. \ref{fig:occlusion}). While on the other hand the technology is cost effective and portable as it requires no additional setup or installation.

Keeping all this in mind, we present a novel pipeline for 3D reconstruction of the human body from cameras installed on VR headsets. Our pipeline includes two modules, one for view translation and one for the 3D reconstruction. In the first module, the egocentric (first person) images from the VR cameras are translated into third person full body views of the subject. This module is trained in an adversarial manner and uses an architecture similar to that proposed by \cite{pix2pix2017}. The output from the first module is then sent through our second module which gives us the reconstructed mesh. We use the SMPL \cite{SMPL:2015} body model which makes the output mesh very easy to import into 3D modelling software and game engines, and can further be animated as well. Our method does not require any pre-scans of the users body and can adapt to new users with varying body shapes without any tweaking of the model. We also talk about a synthetic dataset that we made for our research.


\begin{figure*}[t]
	\centering
	\includegraphics[width=\linewidth]{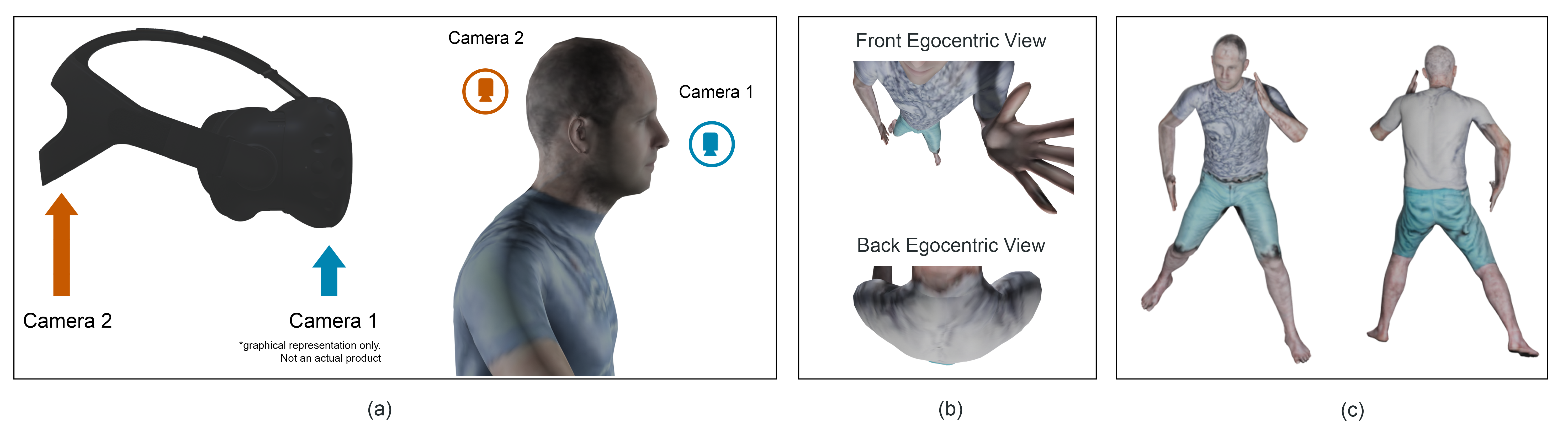}
	\caption{(a)The placement of the cameras on the VR headset. There is a camera in the front pointing downwards capturing the front and another camera on the back capturing the back. (b) The views from the front and back VR cameras. You can see that they are severely distorted due to perspective. (c) The third person view of the back and front body}\label{fig:system}
\end{figure*}

\section{Related Work}

\subsection{Non egocentric 3D reconstruction}
For our work, we borrow some knowledge from the classic methods that do 3D reconstruction from images. Methods for 3D reconstruction of static scenes include simultaneous localization and mapping \cite{LSD-SLAM}\cite{ORB-SLAM}, using depth sensors along with cameras \cite{Izadi2011KinectFusionR3} \cite{matterport} \cite{Newcombe2011KinectFusionRD} and using stereo vision \cite{Scharstein2001ATA}. These works work well for static scenes only. 

For mobile objects, such as humans and cars, motion capture systems \cite{Aguiar2007MarkerlessDM} \cite{Starck2007SurfaceCF} \cite{Vlasic2008ArticulatedMA} have long been in use. The paper by Silva \etal \cite{Silva2019UsingLF} uses deep learning to classify human motion from motion capture files. They use a long short-term memory network (LSTM)\cite{Hochreiter1997LongSM} trained to recognize action on a simplified ontology of basic actions like walking, running or jumping. 
Motion capture based tracking usually require tracking markers, depth based sensors and cameras, pre-scanned body models and dedicated environments and lighting. The equipment used are usually very high in cost and low in portability. This prevents them to be used in virtual reality based applications. 
3D reconstruction of humans from one or more RGB images has been achieved in many ways.
Haberann \etal\cite{Habermann2019LiveCapRH} propose a monocular real-time human performance capture. Their method, although promising, requires the camera to be at a distance from the subject which reduces the area of use. For constructing the mesh and the texture, they require a template model that they create using multi-angle images of the actor taken in a static pose. This makes their model non-adaptive and person specific.
Parametric body models have also seen remarkable success in reconstructing the human body from RGB images. The work by Anguelov \etal \cite{SCAPE} represent body shape and pose-dependent shape in terms of triangle deformations and by training on body scans containing unique structure and poses, they learn a statistical model for the shape variation. This was enhanced further by Chen \etal \cite{Chen2013DeformableMF} when they also took into account the deformation of the clothing into their parameters. The SMPL model proposed by Loper \etal \cite{SMPL:2015} uses a similar set of parameters for body shape and pose and gives a much more physically accurate model of the body. 
Using SMPL and gender specific models, the work by Pavlakos \etal\cite{Pavlakos2019ExpressiveBC} focuses on hands and expression alongside the pose of the body.

The work by Bogo \etal \cite{Bogo:ECCV:2016} uses an optimization based method to recover the parameters for SMPL from the output of CNN based keypoint detector. Kanazawa \etal \cite{HMR} use adversarial training on unpaired dataset of static 3D human models and poses. Their model works in a frame-by-frame approach. Extended to videos, the work by Kocabas \etal \cite{kocabas2019vibe} uses a recurrent architecture with an adversarial objective for inferring the shape and pose parameters.
The approach for creating neural avatars by Shysheya \etal\cite{Shysheya2019TexturedNA} splits the body into multiple parts and matches each pixel to these parts. Using the matched pixels the texture is generated. Their model relies on pose estimation and suffers significantly when there is error in the pose estimation.

\subsection{Egocentric 3D reconstruction of the body}

Next we look at methods for 3D reconstruction from Egocentric or wearable cameras. Reconstructing faces from wearable cameras is a unique and challenging problem because the object we want to track is distorted and largely occluded by the headset the camera is mounted on. Hence these techniques attempt to capture the full face through multiple sensors installed inside and around the headset.

To handle occlusions while having the minimum numbers of cameras, Wei \etal \cite{Wei2019VRFA} used a multiview translation method in which they train a network that lets them augment additional views. Inspired from this we train a network which augments the third person view from the egocentric images.




Rhodin \etal \cite{Rhodin2016EgoCapEM} use two fisheye cameras attached to a helmet to predict the pose of the user in real time. Extending this idea further, Xu \etal \cite{xu2019mo2cap2} use only a single camera attached to a baseball cap and they are able to predict the 3D pose of the user in real time.

The work done by Cha \etal\cite{Cha2018TowardsFM} is able to capture the motion of the user from head-worn cameras. They use a pre-scanned model of the user and transfer the motion of the user onto it. They are also able to reconstruct the environment and localize the user's position within it. Using both audio and video for reconstructing and re-tagging the faces their work shows promising results. Their work however is very user specific. For a new user, a new pre-scan using multi-angled images of the person is needed and the pose estimation model has to be trained again for the new user. This limits the feasibility of their work since capturing the pre-scans and dataset for pose estimation require a separate dedicated setup.


\section{System Overview}

Our proposed system consists of a wearable VR headset with two cameras installed on it, one of the front and one on the back. To make it easier to visualize the setup, we have shown the camera placement and the simulated views in Fig. \ref{fig:system}. 

The camera on the front of the VR headset captures the view of the body from the front and is more prone to occlusions (such as the hands blocking the legs) than the camera on the back. Previous works such as that by Rhodin \etal\cite{Rhodin2016EgoCapEM} use two cameras for perceiving the depth of the image using stereoscopic vision. Our method on the other hand does not require the depth information. We use two cameras to be able to understand the body posture in scenarios where a single camera on the front won’t be enough. For example if the person is looking slightly upwards or if the person is bending backwards, the camera on the front barely captures the body but is pointed more towards the environment, but in the same situation the camera on the back of the VR headset gets a complete view of the body and is used by our model to infer the body structure and appearance. We assume that the user is in a solid colored room and the body is easily differentiable from the background. 

Since there doesn’t publicly exist a dataset specifically for VR based cameras that we could use for our work, we created a simulated dataset. We will elaborate more on that in a separate section. 

\begin{figure*}[t]
	\centering
	\includegraphics[width=\linewidth]{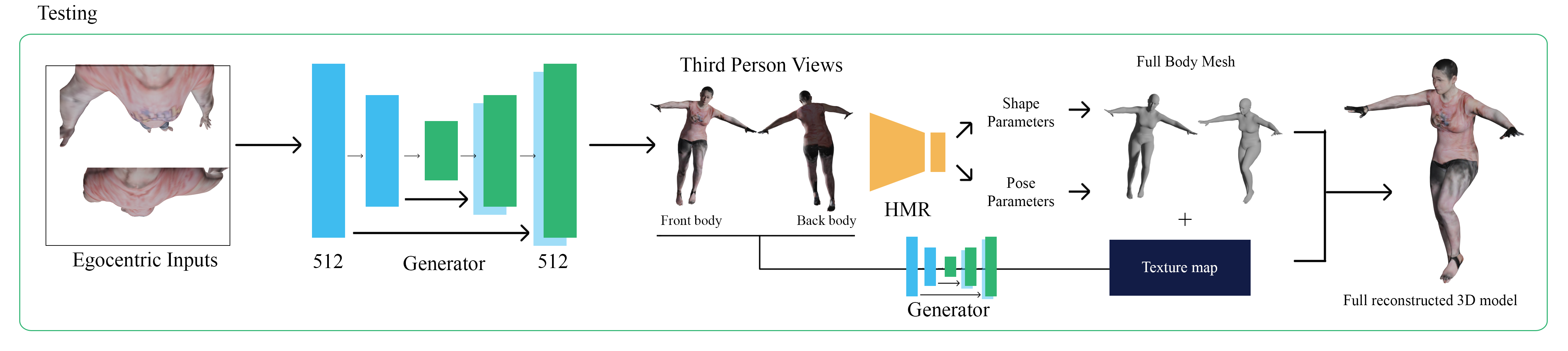}
	\caption{Pipeline for reconstructing the 3D model of the human body from egocentric images. The inputs from the egocentric camera are first sent through an image translation network which translates the egocentric views into third person full body views. Then the translated view is sent through the 3D reconstruction module which outputs the shape and pose parameters for the SMPL model and the texture maps are obtained using the texture generation model. The reconstructed 3D mesh can be animated and viewed from novel viewpoints}  \label{fig:pipeline}
\end{figure*}

An overview of our reconstruction pipeline can be understood from Fig. \ref{fig:pipeline}. The whole process is divided into two steps: view translation and 3D reconstruction. In the view translation step, the input to the network is a vertically stacked version of the images from the front and back VR cameras (first person view) and the output is an image containing full body front and back view (third person view). The view in the input images are largely affected by perspective and occlusions and this step allows us to translate them into a more comprehensible form for the 3D reconstruction. The output from the view translation step goes as input into the second module for the 3D reconstruction and the 3D model of the person is obtained as the final output. A subset of the 3D reconstruction module is a texture generation model which generates easy-to-apply texture maps for the generated mesh.  

The view translation step allows us to take advantage of the existing datasets for 3D human scans and their corresponding renders such as \cite{h36m_pami} \cite{varol17_surreal} \cite{mixamo} and the existing state of the art methods for image to 3D reconstruction of the human body such as \cite{kocabas2019vibe} \cite{SMPL:2015} \cite{pifuSHNMKL19}.

Reconstruction of the face and the facial expressions accurately is out of the scope of our system. It has already been achieved by \cite{Wei2019VRFA} and can be used with our system as an extension. However we included the facial structure of the simulated characters in our dataset as it might prove to be useful for further research. Since the face is barely in the input images, the model will try to reconstruct the face using the existing data in its latent space.

\section{Dataset}

At the time of performing this, there didn’t exist a dataset that we could’ve directly make use of for the work we are performing. The datasets released by \cite{Rhodin2016EgoCapEM} and \cite{xu2019mo2cap2} consisted of a large corpus of egocentric fish-eye images along with the detailed pose annotations for each image, but it gave no information about the occluded body parts and the body parts that were not in the field of view of the camera. Hence for our problem, we curated our own dataset.

\begin{figure}[t]
	\centering
	\includegraphics[width=\linewidth]{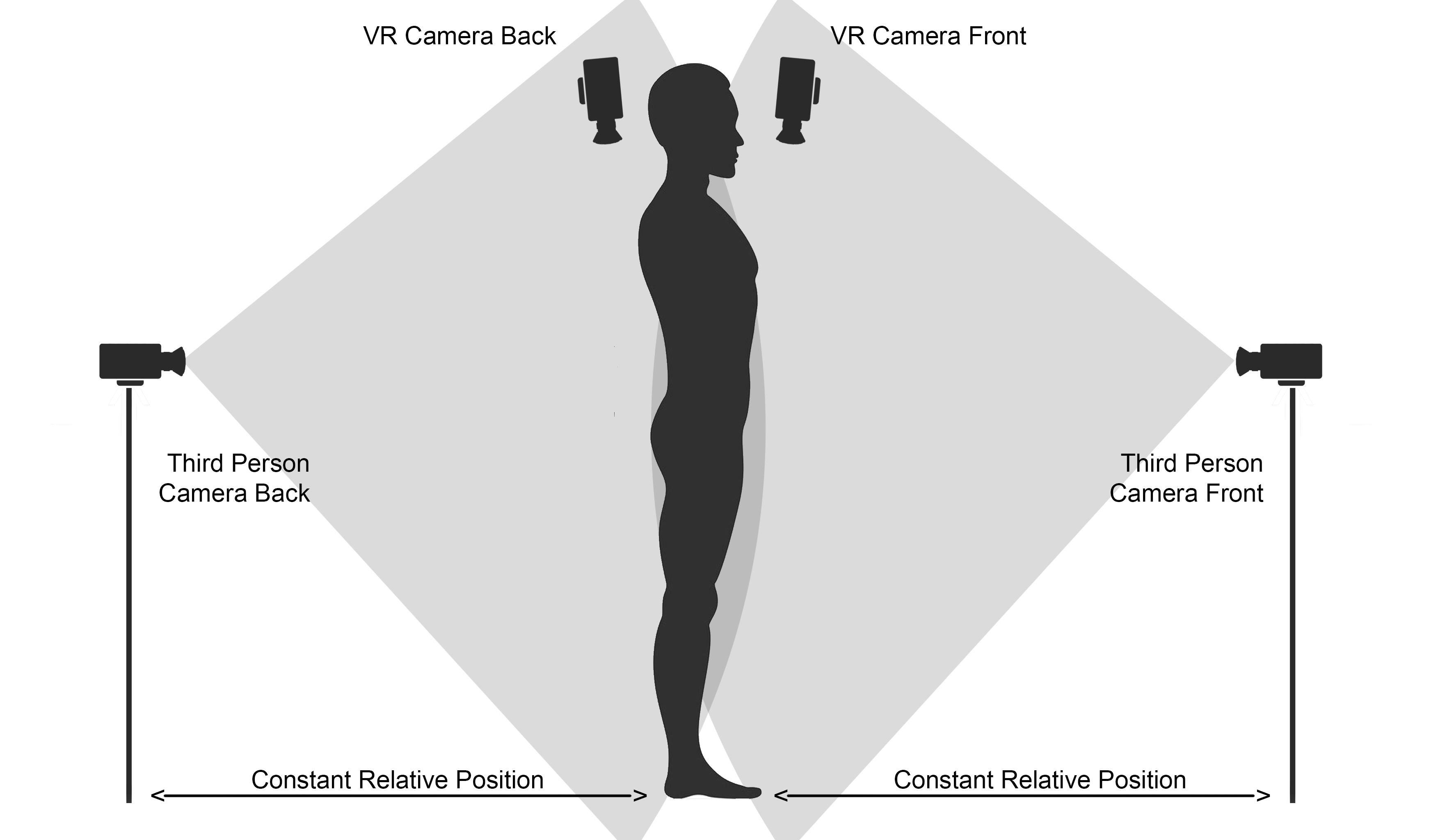}
	\caption{Camera setup for dataset collection. Two cameras are placed on the VR headset and two cameras are placed at a distance from the user to capture the front and back third person views.}
	\label{fig:camera setup}
\end{figure}

The guidelines for making the dataset were simple. For each pair of front-back egocentric images, there should be a corresponding pair of front-back full body third person views of the person. This would allow the deep learning model to learn to cater to occlusions and situations where a body part is out of the field of the view of the camera. Fig. \ref{fig:camera setup} shows how such a setup would look like. The biggest obstacle of creating such a dataset was to get the front-back third person views for each frame. We required a dataset of our subjects performing various activities and every-time the subject would rotate and move around, we would have to move the camera around him so that their relative motion is null. Only this would allow us to get the accurate front-back third person views as the subject performs various activities. To create such a system was a mammoth task and impractical.

So we decided to solve this problem by creating a large synthetic dataset that is tailored to suit our system. We used the SMPL model \cite{SMPL:2015} and attached 4 cameras to each model according to the setup in Fig. \ref{fig:camera setup}. One camera is attached in front of the head at a distance approximately where the bottom of a VR headset would be and another camera is attached behind the head approximately where the strap of the VR headset would be. Two more cameras for the third person views are placed in front and back of the character. These two cameras are attached to the hip joint of the rigged skeleton of the model. This allows the cameras to move relative to the subject as it performs various activities. The subject could be back-flipping or samba dancing and the cameras will always capture the accurate front-back third person views. For texturing the mesh, we used the texture maps released by \cite{varol17_surreal} and applied them to the model. The multiple shape parameters of the SMPL model allowed us to create several variations in the body shape of the subject. We used \cite{mixamo} to rig and animate the models. We made the models perform over 50 distinct activities including boxing, jumping, dancing, walking (activities that might be performed in VR games or in virtual meetings) and we rendered a total of 50,000 frames and for each frame there is a pair of front-back egocentric views and a pair of front-back third person views.

The background for each frame is a solid color which allows for manually augmenting the backgrounds as per the need of the research.

\section{Image to Image translation}

\begin{figure*}[h]
	\centering
	\includegraphics[width=\linewidth]{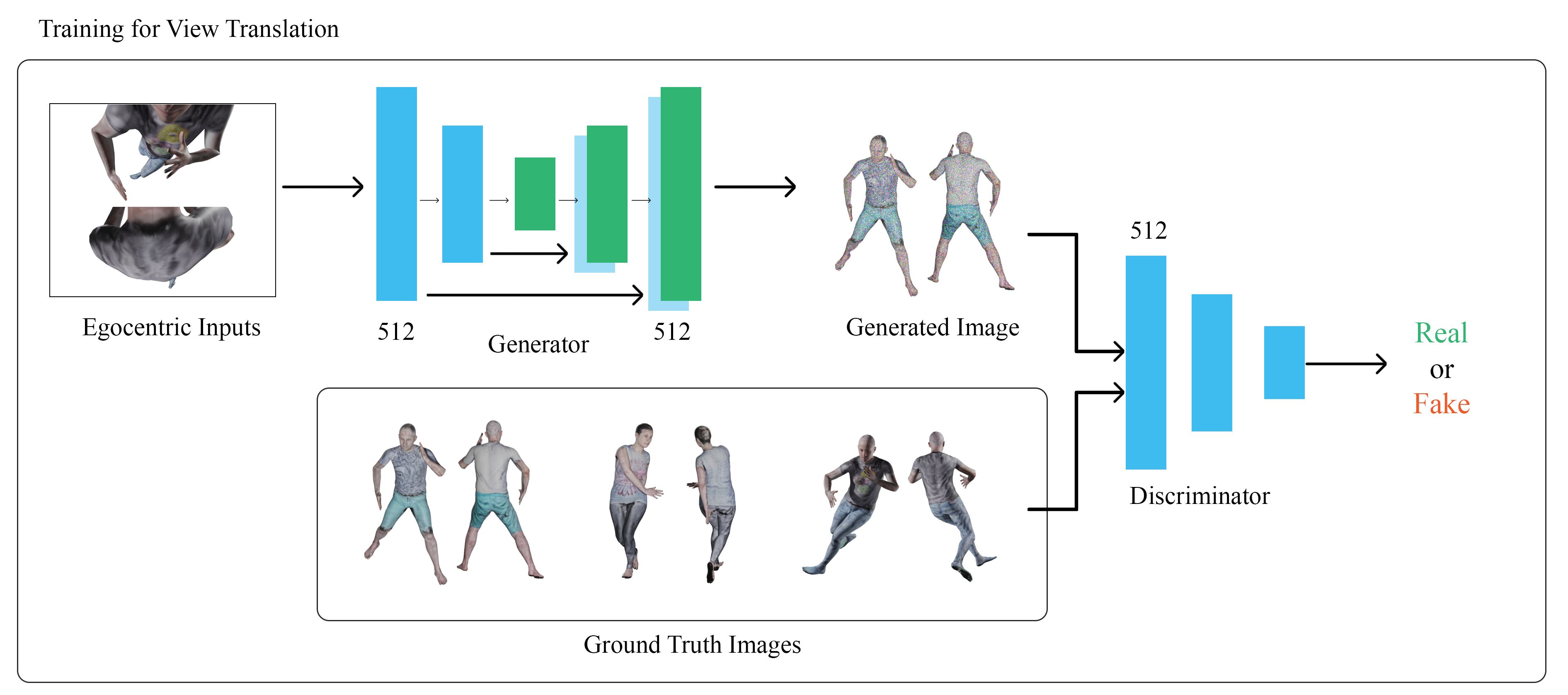}
	\caption{The training of the view translation module. The generator takes as input the egocentric image and tries to generate the third person views for it. The generated image is then fed to the discriminator which classifies it as real or fake. They are both trained simultaneously until the generator starts outputting realistic third person images that correspond well with the egocentric images.}  \label{fig:pipeline train}
\end{figure*}

The first step to reconstruct the user’s body is to obtain third person full body views from the egocentric views. This step is crucial as it allows the system to be able to understand the distortion of the image due to perspective and to infer the occluded and non-visible regions of the body before it is reconstructed in 3D.

The main aim of this step is to generate an output of type \textit{y} given an input of type \textit{x}. Generative Adversarial Networks (GANs) \cite{NIPS2014_5423} have performed remarkably well in the deep learning based generative area of study. Their architecture consists of two models, a generator \textit{G} and a discriminator \textit{D}. The job of the generator is to generate realistic examples relative to the training dataset and the job of the discriminator is to classify an image as realistic or fake. \textit{G} and \textit{D} are both trained together in a two-player min-max situation. But GANs are only effective in generative image synthesis applications if we need to generate new examples of images. We have no control on the data being generated. To be able to control the outputs and to make use of additional information, such as class labels, or in our case an input image of type \textit{x} that we want to be translated into an image of type \textit{y}, we use an extension of GANs called Conditional GANs \cite{Mirza2014ConditionalGA}.

In conditional GANs, the generator \textit{G} learns to generate fake samples with a condition instead of unknown noise distribution. The final objective of a conditional GAN looks like
\begin{equation}
\begin{aligned}
\mathcal{L}_{c G A N}(G, D)=& \mathbb{E}_{x, y}[\log D(x, y)]+\\
& \mathbb{E}_{x, z}[\log (1-D(x, G(x, z))]
\end{aligned}\label{eq1}
\end{equation}

%

\begin{figure}[t]
	\centering
	\includegraphics[width=\linewidth]{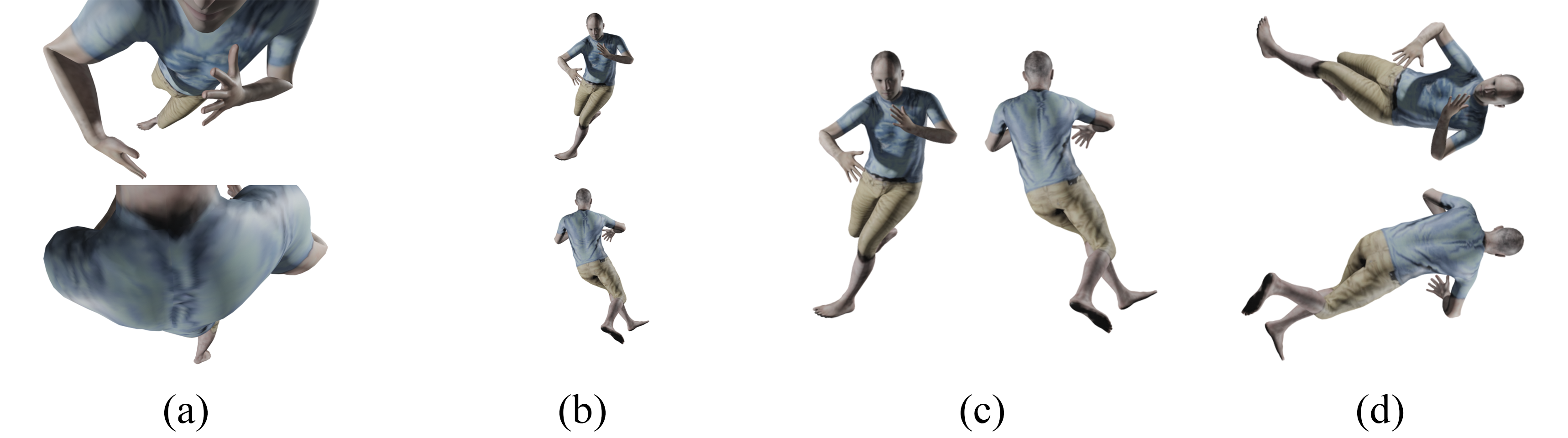}
	\caption{The different ways to arrange the ground truth image for feeding into the view translation module. The first column shows the input from the VR headset camera. The second, third and fourth column show the different ways to arrange the front and back third person images. 
	In (b), the correlation between the input and the ground truth is the highest since the front and back egocentric images are put right next to the front and back third person images respectively. But this results in lower quality images. In (c), the third person views are stacked next to each other and are scaled up but the top bottom correlation is lost. In (d), the stacked views from (c) are rotated clockwise establishing the correlation while keeping the image quality high.
	}\label{fig:ground truth}
\end{figure}

Our problem is of the type image to image translation and there has been some remarkable progress in this field when combined with conditional GANs. Conditional GANs have been used to achieve tasks like colorization of black and white images by Isola \etal \cite{zhang2016colorful} ,  future frame prediction \cite{futureframe}, image prediction from normal maps \cite{normal_map} etc. The work that we decided to use for our research by Isola \etal \cite{pix2pix2017} consists of a very general image to image translation architecture which has been used to several applications by researchers later such as pose transfer\cite{Chan2019EverybodyDN} , edges to realistic images, simulation to reality etc. They also incorporate a convolutional PatchGAN classifier for the discriminator which allows the structure to penalize at the scale of image patches. So instead of trying to check whether the image as whole is real or not, the PatchGAN checks whether each N x N patch in the image fed to the discriminator is real or not. Then the predictions by the discriminator for all patches are averaged and given out as the final output.

Along with the cGAN loss in \eqref{eq1}, they also use a traditional L1 loss. This forces the generator to generate images near the ground truth output in an L1 sense while also trying fool the discriminator into believing the generated images are real.
\begin{equation}
\begin{aligned}
\mathcal{L}_{L 1}(G)=\mathbb{E}_{x, y, z}\left[\|y-G(x, z)\|_{1}\right]
\end{aligned}\label{eq2}
\end{equation}
This results in their final objective function as,

\begin{equation}
\begin{aligned}
G^{*}=\arg \min _{G} \max _{D} \mathcal{L}_{c G A N}(G, D)+\lambda \mathcal{L}_{L 1}(G)
\end{aligned}\label{eq3}
\end{equation}

Apart from the PatchGAN, their generator network uses a U-Net \cite{Ronneberger2015UNetCN} style architecture which allows them to establish a better relation between the input and output images that have the same low level structure such as in image colorization and simulation to reality. In our case though this feature is not as useful since our input images and output images are considerably different. Though this doesn’t prove to be a disadvantage either as the network without the U-Net architecture gave similar results to the original Pix2Pix network. 

The training of the network is straightforward; the model is fed with the vertically stacked front-back pair of images from the VR camera as input and the combined third person views for ground truth. The ground truth could be fed in three different ways as seen in the Fig. \ref{fig:ground truth}. Fig. \ref{fig:ground truth}(b)  might seem like a better option at first since we can make a direct correlation between the input and output visually since the egocentric front image is aligned with the third person from image and so on for the back images, but this gave low quality results as a huge portion of the image was blank and wasted. In Fig. \ref{fig:ground truth}(d) the third person views are scale up but there is no direct correspondence between the first person and third person views. In Fig. \ref{fig:ground truth}(c), the third person views are not only scaled up and also aligned horizontally with their corresponding first person views. The output images are rotated clockwise, which isn't really an issue as long as the model is able to learn the correspondence well. The experiment results for each orientation can be seen in the Results section.

\begin{figure*}[t]
	\centering
	\includegraphics[width=\linewidth]{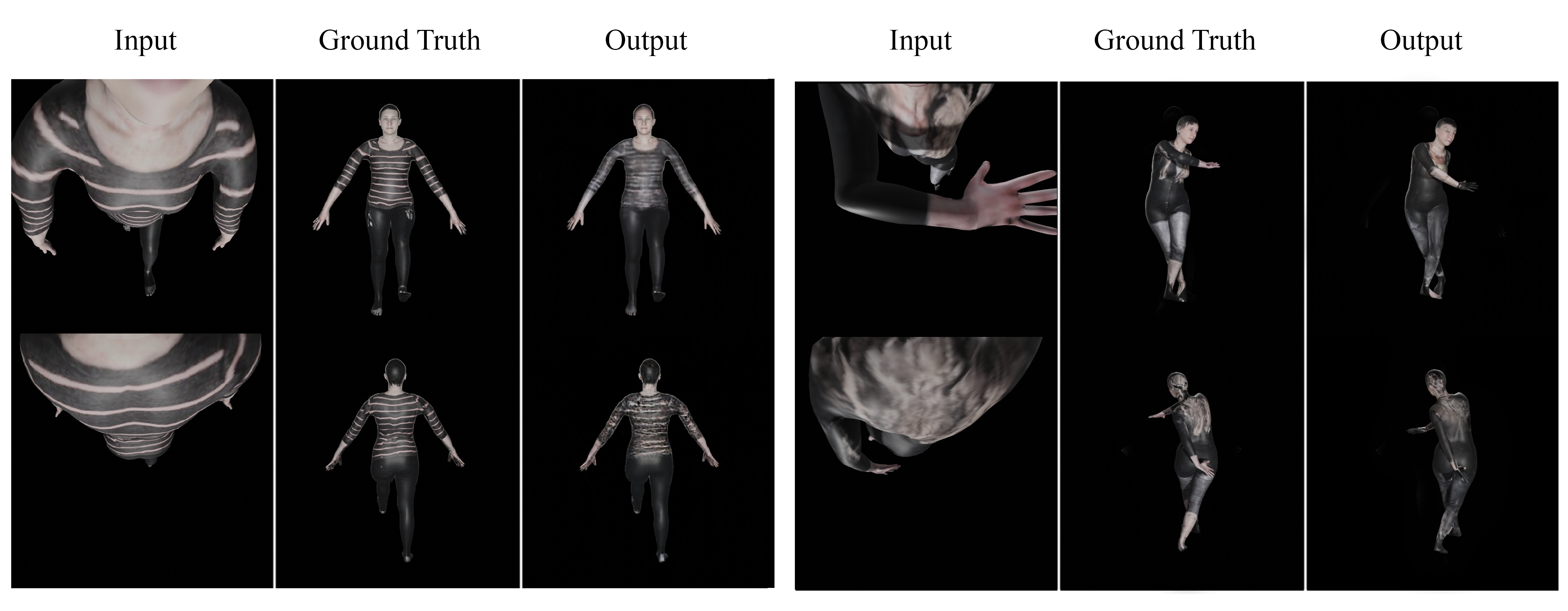}
	\caption{Example results for the view translation step. The model is not only able to infer the body shape and the color of the person's clothes, but also minor details such as stripes and patterns.}
	\label{fig:i2igt}
\end{figure*}

\section{3D Reconstruction}
Once we have the output from the view translation step, we are ready to reconstruct the 3D body from it. For reconstruction of the body we chose the generative human body model SMPL\cite{Bogo:ECCV:2016} which is a realistic human body model and its body shape and the pose can be controlled by tweaking shape and pose parameters. Since the rig of the SMPL model has 23 joints, any pose \( \theta \) can be defined with \( |\vec{\theta}|=3 \times 23+3=72 \) parameters; i.e. 3 for each part plus 3 for the root orientation. Similarly the body shape \( \beta \) has 10 parameters.

Using the SMPL model will also allow us to animate the reconstructed model. The basic idea here is to fit the SMPL model to our subject’s body images by accurately estimating the parameters required for its body shape.This has been achieved earlier by Bogo \etal \cite{Bogo:ECCV:2016} which fits the SMPL model to the output of a CNN keypoint detector. Other methods \cite{holopose}\cite{neural}\cite{pavlakos2018humanshape} include training neural networks which use the pixels to directly regress the parameters. More recent approaches such as \cite{HMR}\cite{kocabas2019vibe} use adversarial objectives to infer the parameter for SMPL. We follow the adversarial approach for our work as well. The HMR \cite{HMR} model works best for still images while VIBE \cite{kocabas2019vibe} goes a step further and employs a recurrent architecture with the adversarial objective which allows it to work remarkably well with videos. 

For the scope of this research, implementing and applying HMR was more feasible and efficient since we only needed to know the body shape parameters from a single image. Along with the pose \( \theta \) and the body shape \( \beta \), they also use a weak-perspective camera model and solve for global rotation \( R \in \mathbb{R}^{3 \times 3} \) in axis angle representation, scale \( s \in \mathbb{R} \) and the translation \( t \in \mathbb{R}^{2} \). This finally leaves them with a set of parameters to represent the reconstruction of any 3D human body which can be expressed as \( \Theta=\{\boldsymbol{\theta}, \boldsymbol{\beta}, R, t, s\} \). \( \Theta \) is an 85 dimensional vector.

The objective function that they use to train their final model is given by,
\begin{equation}
L=\lambda\left(L_{\text {reproj }}+\mathbb{1} L_{3 \mathrm{D}}\right)+L_{\text {adv }} \label{eqHMR}
\end{equation}
where \( L_{\text {reproj }} \)  is the joint reprojection error, \( L_{3 \mathrm{D}} \) is the 3D Error, \( L_{\text {adv }} \) is the adversarial loss,  \( \lambda \) controls the relative importance of each objective and \( \mathbb{1} \) is a function that is 1 only if the 3D ground truth is available for the image, otherwise it is 0. 

The input to this module is the third person view of the front body obtained in view translation and the output of the module is the set of parameters required to reconstruct the 3D body. By applying the obtained body shape parameters to the SMPL model we successfully reconstruct the 3D rigged mesh of the body and by applying the obtained pose parameters we can transform the model according to the pose of the person in the input image.\

For the texture we trained another image-to-image translation model which takes as input the third person views generated in step 1 and outputs the corresponding texture maps. These texture maps can be used as is with the generated model and require no tweaking of the UV coordinates.

\begin{table}[t]
	\caption{Evaluating View Translation with different ground truth arrangements}
	\begin{center}
		\begin{tabular}{|c|c|c|c|}
			\hline
			\cline{2-4} 
			\textbf{Metric} & \textbf{\textit{Method A}}& \textbf{\textit{Method B}}& \textbf{\textit{Method C}} \\
			\hline
			RMSE& 89& 53.2&\textbf{40.1}  \\
			\hline
			SSIM& 0.67&0.72 &\textbf{0.89}  \\
			\hline
		\end{tabular}
		\label{tab1}
	\end{center}
\end{table}

\begin{figure*}[!tb]
	\centering
	\includegraphics[width=\linewidth]{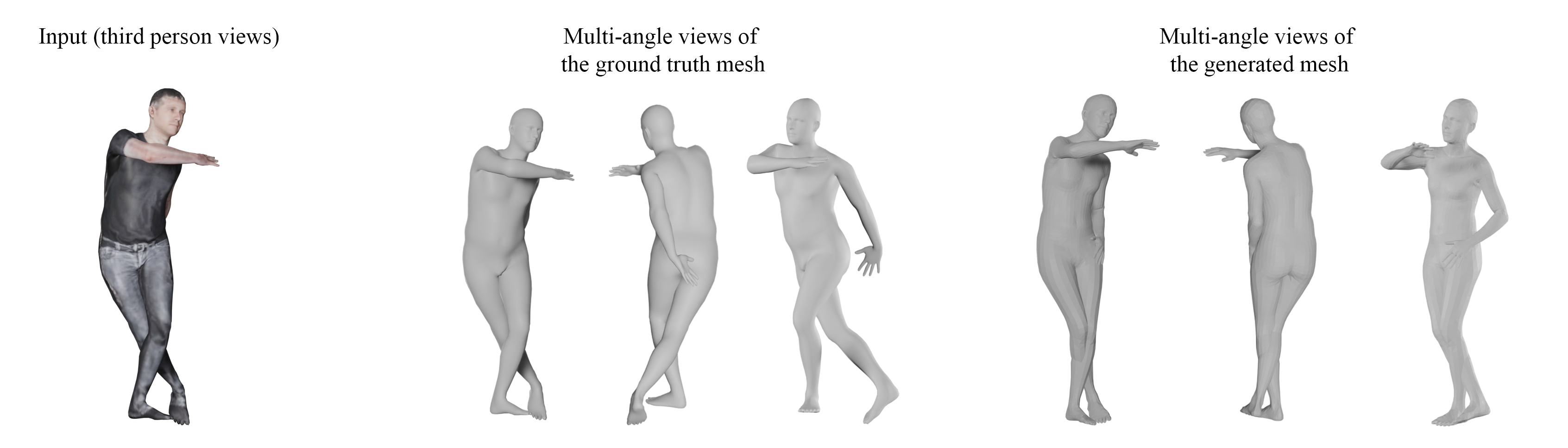}
	\caption{Evaluation of the mesh reconstruction step. The mesh generated from the inferred third person views are compared to the ground truth mesh. Even for such a complex pose consisting of occlusions, the mesh generated is fairly accurate.}
	\label{fig:3D eval1}
\end{figure*}

\begin{figure*}
	\centering
	\includegraphics[width = 0.88\textwidth]{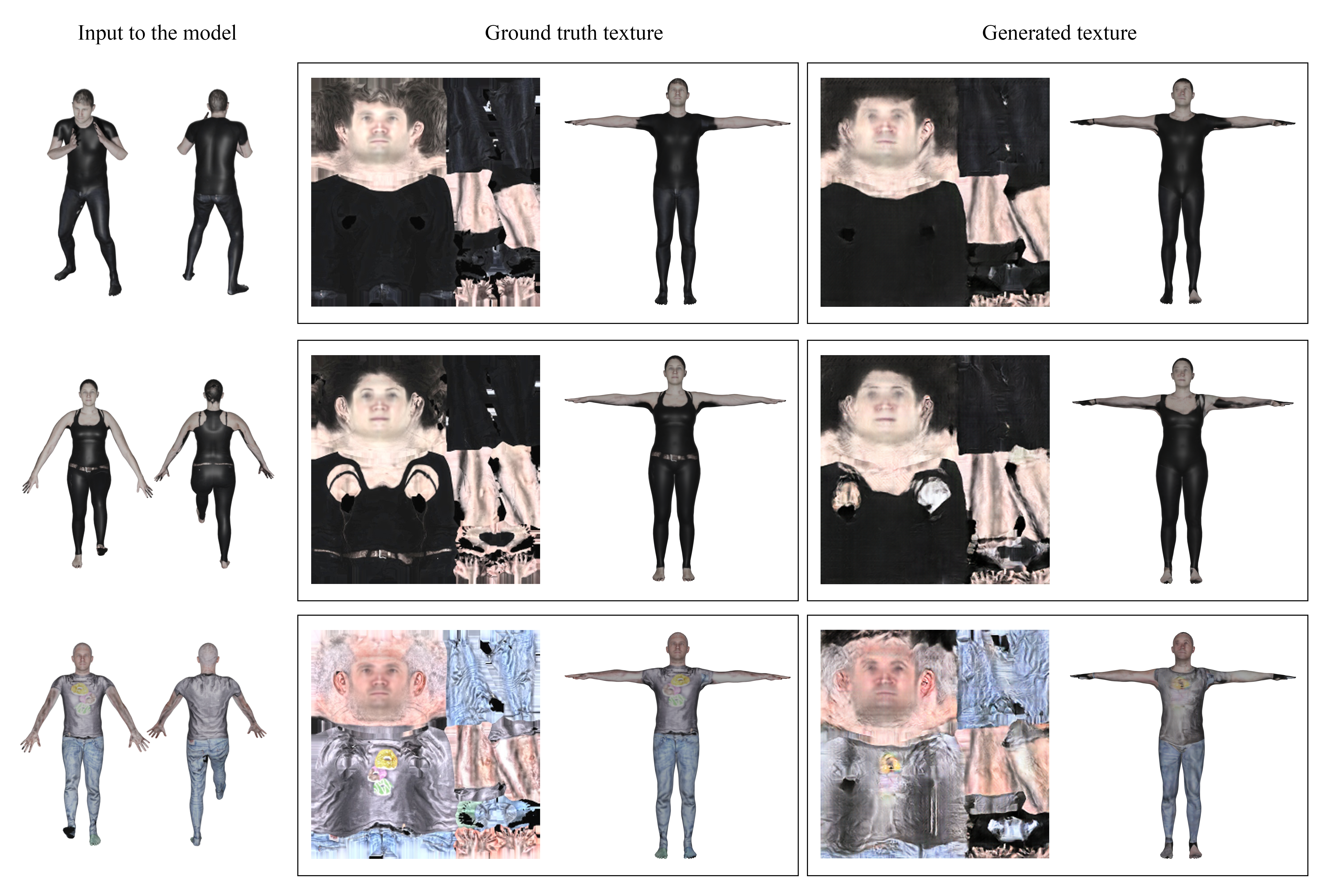}
	\caption{Evaluation of the texture generation model. The left column shows the third person view input images to the model. The middle and the right columns respectively show the ground truth and generated texture maps along with the textured mesh.}
	\label{fig:texture}
\end{figure*}

\begin{figure}[!tb]
	\centering
	\includegraphics[width=\linewidth]{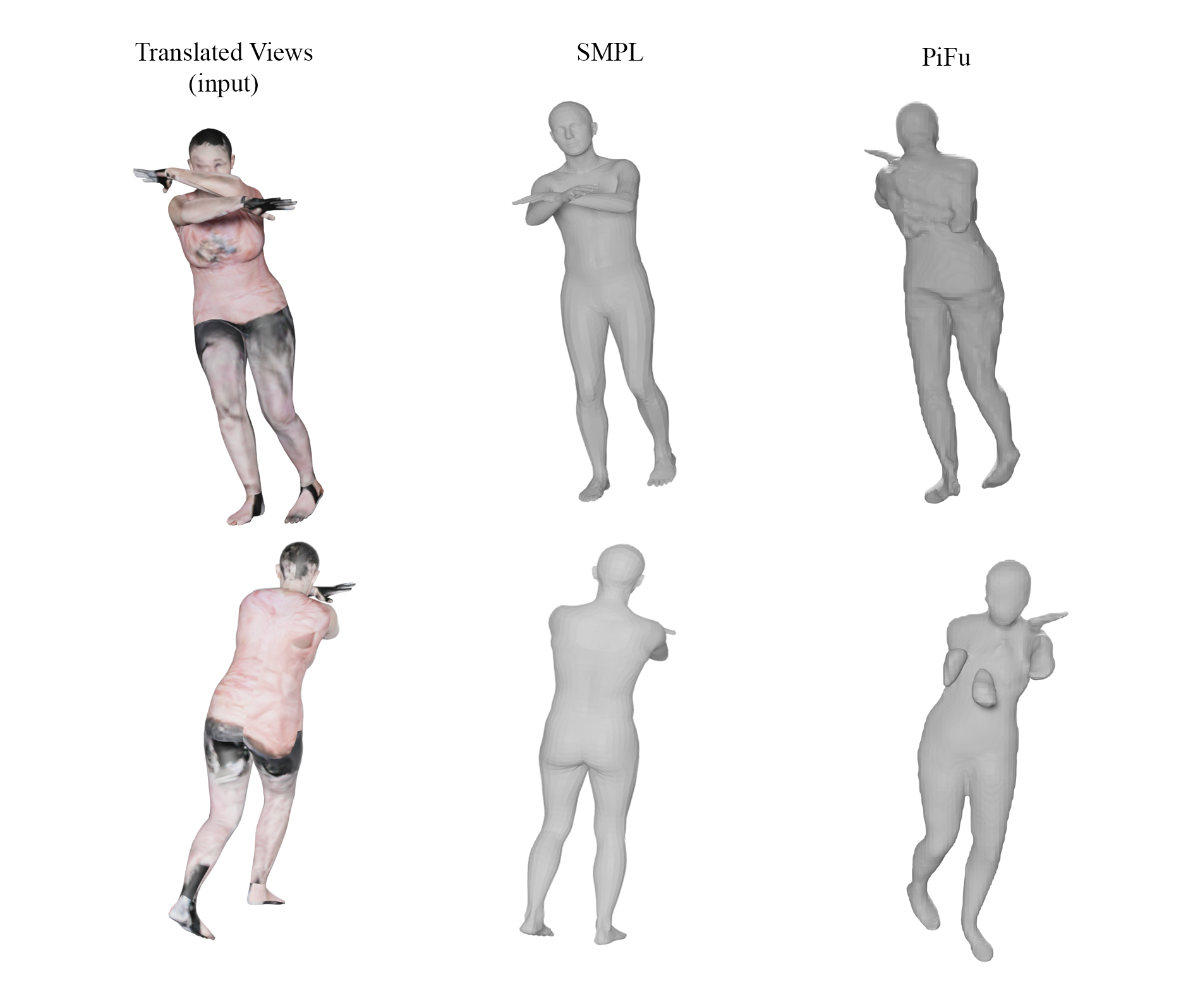}
	\caption{The mesh generated from the SMPL model is compared with the mesh generated using PIFu. PiFu generates a distorted mesh of the body due to occlusions by the hand whereas the mesh recovery for the SMPL model gives fairly accurate results}
	\label{fig:pifuvsours}
\end{figure}

\begin{figure*}[h]
	\centering
	\includegraphics[width=\linewidth]{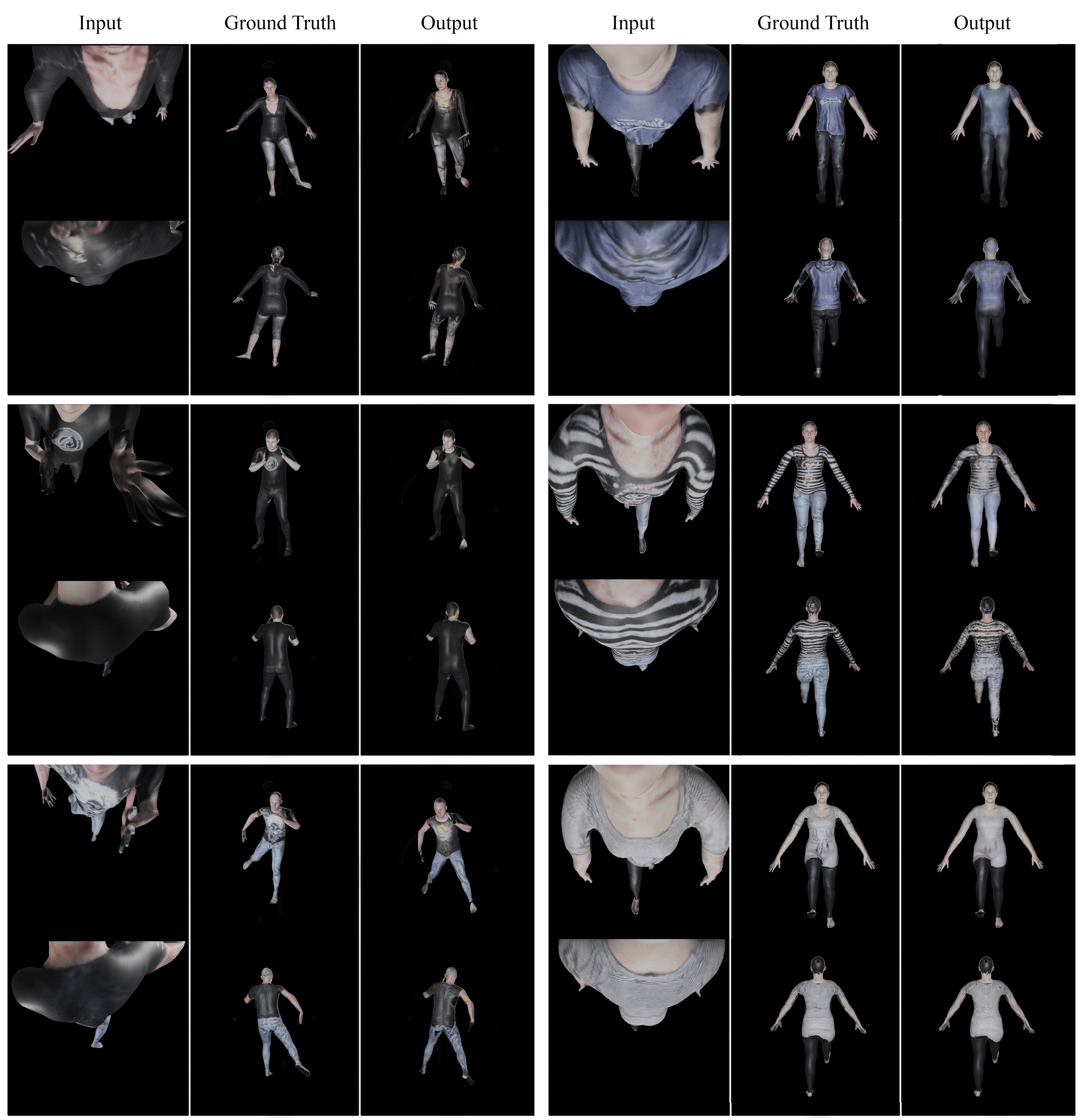}
	\caption{Example results for the view translation module.}
	\label{fig:ego examples}
\end{figure*}

\begin{figure*}[t]
	\centering
	\includegraphics[width=\linewidth]{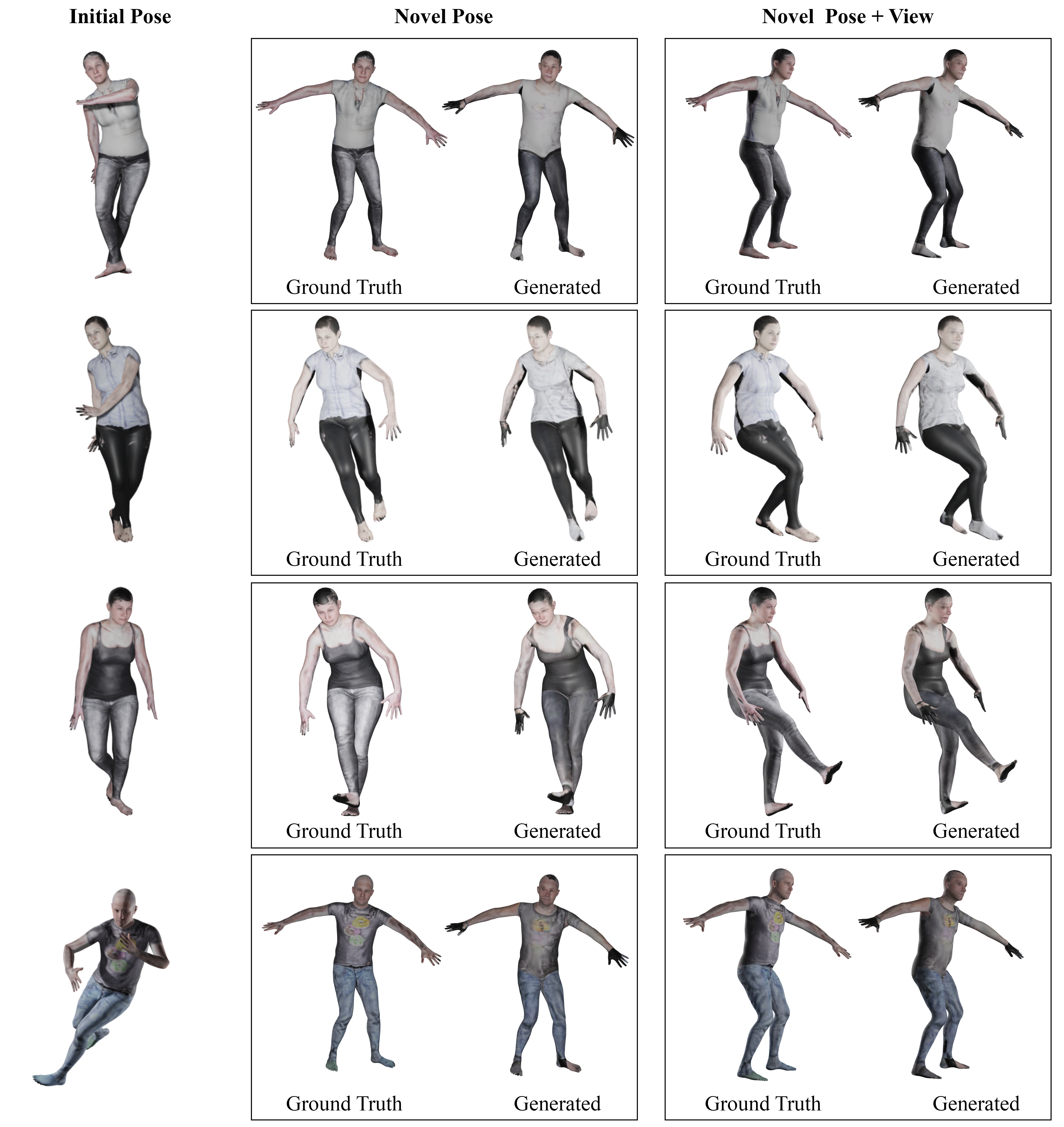}
	\caption{Example results for animation of the mesh. We compare the generated and the ground truth texture in novel poses and view points.}
	\label{fig:novel anim}
\end{figure*}

\section{Results}
In this section we will show and evaluate the results of our view translation and 3D reconstruction pipeline.
\subsection{View Translation}
For view translation, the first experiment that we conducted was to establish the best method to arrange the third person views before feeding it into the model as ground truth. We checked three different methods of arranging them as seen in Fig. \ref{fig:ground truth}. 
We trained a model three times on the same dataset and each time arranging the ground truth images differently. To quantitatively evaluate them, we checked the mean squared error values and the SSIM between the output images and their corresponding ground truth images on a test dataset containing 200 images. We show the average values in table \ref{tab1} where Method A corresponds to Fig. \ref{fig:ground truth}(b), Method B corresponds to Fig. \ref{fig:ground truth}(c), and Method C corresponds to \ref{fig:ground truth}(d). It should be noted however that for evaluating the ground truth for Method A, we cropped and scaled up the region including the subject to match the size of the other two types of ground truth orientations. This is done since having a high quality larger image in the output was a driving factor for this experiment and also the average error between the images of two very small artifacts would be smaller generally.

Next we show the results of the final model on unseen input images in Fig. \ref{fig:i2igt} and Fig. \ref{fig:ego examples}. On comparing the generated results with the ground truth we get the average SSIM value as 0.89 and RMSE value as 40.1. We further use a state of the art pose detector \cite{open-pose} on the generated and ground truth images to quantitatively evaluate the accuracy of the generated pose and we get an average RMSE of 10.21 between the generated joint values and the ground truth joint values.

\subsection{3D reconstruction}
To evaluate the 3D reconstruction model individually, we first input non-generated ground truth third person views to the model and compare the SMPL model generated from it to the actual model for those images. (Fig. \ref{fig:3D eval1}). In the example shown, there is a high level of occlusion since the right hand is not visible at all and it can be seen that the model is fairly accurate even for such a complex pose. 


Furthermore, we compare the performance of the SMPL model that we are using with another state of the art method for reconstruction of the human body from a single RGB image by Saito \etal \cite{pifuSHNMKL19}. Their work PIFu is able to generate the texture and the mesh remarkably well for images of humans, but if we pass generated third person image which has massive occlusions (such as the hands covering the chest) to their model, it can be seen that their model is unable to reconstruct the body realistically whereas using the huamn mesh recovery method for the SMPL model works really well (Fig. \ref{fig:pifuvsours}). Furthermore their generated mesh is not rigged. Since the SMPL approach uses a rigged 3D model, it can be animated later if we import it in a virtual reality game or if we are able to extract the pose from the VR camera, we can simply transform the 3D model using the new pose. We show this in Fig. \ref{fig:novel anim}.

\subsection{Texture map generation}
To evaluate the texture map generation model, we use a real third person view rendered from an actual 3D model as input and compare the generated textures to the ground truth textures (Fig. \ref{fig:texture}). It can be seen that the model is able to infer the colors of the clothes easily and it also tries to adapt to other details such as varying types of sleeves and patterns on the t-shirt.

\begin{table}[!bt]
	\caption{Average inference time for each module on a Tesla K80 GPU}
	\begin{center}
		\begin{tabular}{|c|c|}
			\hline
			\textbf{Model} & \textbf{\textit{Inference Time}} \\
			\hline
			View Translation& 0.6 sec  \\
			\hline
			Parameter estimation for SMPL& 0.12 sec \\
			\hline
			Texture Map Generation& 0.56 sec \\
			\hline
		\end{tabular}
		\label{runtime}
	\end{center}
\end{table}

\subsection{Inference Time}
The inference times for the all the steps performed on a Tesla K80 GPU are shown in Table \ref{runtime}. One thing to note about our work is that since the 3D model is rigged, we have to infer the 3D model only once and for every consequent frame it can simply be animated by transferring the pose of the subject. 

\section{Conclusion}
In this paper we presented a novel pipeline for reconstruction of the human body from egocentric cameras. Instead of directly reconstructing the 3D model from egocentric views, we first train a model that can translate the egocentric views into third person full body views. We use one camera on the front and one on the back of the VR headset which allows us to get a better sense of the overall body structure and cater to occlusions by inferring a third person view of the body. From the third person views we estimate the shape and pose parameters for the SMPL model and the corresponding texture map which can be applied to the mesh as is. We further show the reconstructed 3D model from novel view points and in novel poses and compare them to the actual 3D model. The reconstructed model can easily be imported into any 3D modeling software and game engines and can be further animated. Our work can be incorporated with egocentric pose estimation and be animated in real time as well. The mesh will only be generated once in the starting and for every consequent frame only the pose will be estimated and applied to the mesh. This will be useful for several applications such as virtual meetings and conferences, interactive VR games, walk-in movies, remote training and interactive TV shows.


%





\ifCLASSOPTIONcaptionsoff
  \newpage
\fi



%

\clearpage
\bibliographystyle{IEEEtran}
\bibliography{bare_jrnl}




%








\end{document}